\documentclass[letterpaper]{article} 
\usepackage{aaai24}  
\usepackage{times}  
\usepackage{helvet}  
\usepackage{courier}  
\usepackage[hyphens]{url}  
\usepackage{graphicx} 
\usepackage{natbib}  
\usepackage{caption} 
\usepackage{algorithm}
\usepackage{algorithmic}

\usepackage{newfloat}
\usepackage{listings}
\DeclareCaptionStyle{ruled}{labelfont=normalfont,labelsep=colon,strut=off} 
\floatstyle{ruled}
\newfloat{listing}{tb}{lst}{}
\floatname{listing}{Listing}
\title{DeepCalliFont: Few-shot Chinese Calligraphy Font Synthesis by Integrating Dual-modality Generative Models}
\author{
    Yitian Liu, Zhouhui Lian\textsuperscript{\thanks{Corresponding author.}}
}
\affiliations{
	 Wangxuan Institute of Computer Technology, Peking University, Beijing, P.R. China\\
	\{lsflyt, lianzhouhui\}@pku.edu.cn
}

\usepackage{bibentry}
\usepackage{subfigure}
\usepackage{booktabs}
\usepackage{tabularx}
\usepackage{ragged2e}
\newcolumntype{M}{>{\Centering\hangafter=1\hangindent=0em}X}
\newcolumntype{N}{>{\Centering}X}
\usepackage{amsmath}
\usepackage{amssymb}

\begin{document}

\maketitle

\begin{abstract}
Few-shot font generation, especially for Chinese calligraphy fonts, is a challenging and ongoing problem. With the help of prior knowledge that is mainly based on glyph consistency assumptions, some recently proposed methods can synthesize high-quality Chinese glyph images. However, glyphs in calligraphy font styles often do not meet these assumptions. To address this problem, we propose a novel model, DeepCalliFont, for few-shot Chinese calligraphy font synthesis by integrating dual-modality generative models. Specifically, the proposed model consists of image synthesis and sequence generation branches, generating consistent results via a dual-modality representation learning strategy. The two modalities (i.e., glyph images and writing sequences) are properly integrated using a feature recombination module and a rasterization loss function. Furthermore, a new pre-training strategy is adopted to improve the performance by exploiting large amounts of uni-modality data. Both qualitative and quantitative experiments have been conducted to demonstrate the superiority of our method to other state-of-the-art approaches in the task of few-shot Chinese calligraphy font synthesis. The source code can be found at https://github.com/lsflyt-pku/DeepCalliFont.
\end{abstract}

\section{Introduction}

Manually designing a Chinese font is time-consuming and costly, mainly because there exist huge amounts of Chinese characters. For example, the latest official character set GB18030-2022 contains 87,887 Chinese glyphs, most of which possess delicate shapes and complex structures. In the last decade, large numbers of font synthesis methods~\cite{tian2017zi2zi, jiang2017dcfont, lian2018easyfont} have been proposed aiming to automatically generate the complete Chinese font library from a small set of glyphs written or designed by a user. However, existing approaches still have the following two main drawbacks: 1) the number of characters required for training is still large; 2) the quality of machine-generated Chinese fonts is still unsatisfactory for many special font styles, especially cursive Calligraphy styles. 

\begin{figure}[tp]
    \centering
    \subfigure[Assumption 1: the same characters in different fonts have the same components and stroke order.]{\includegraphics[width=0.48\columnwidth]{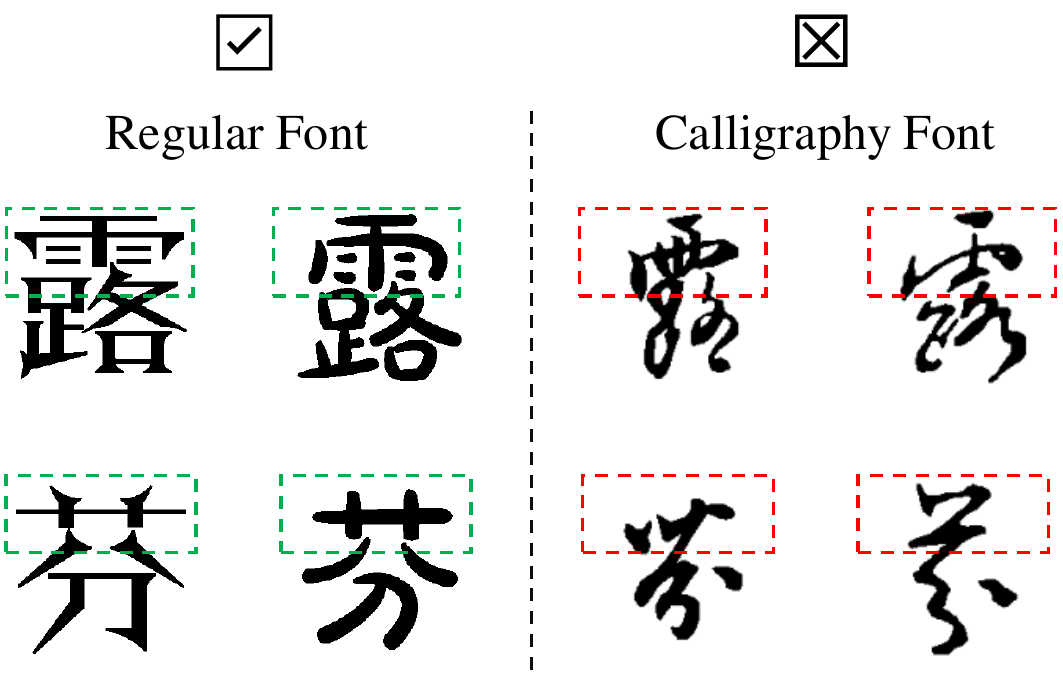}}
    \hfill
    \subfigure[Assumption 2: the same components in different characters but the same font should be identical or similar.]{\includegraphics[width=0.48\columnwidth]{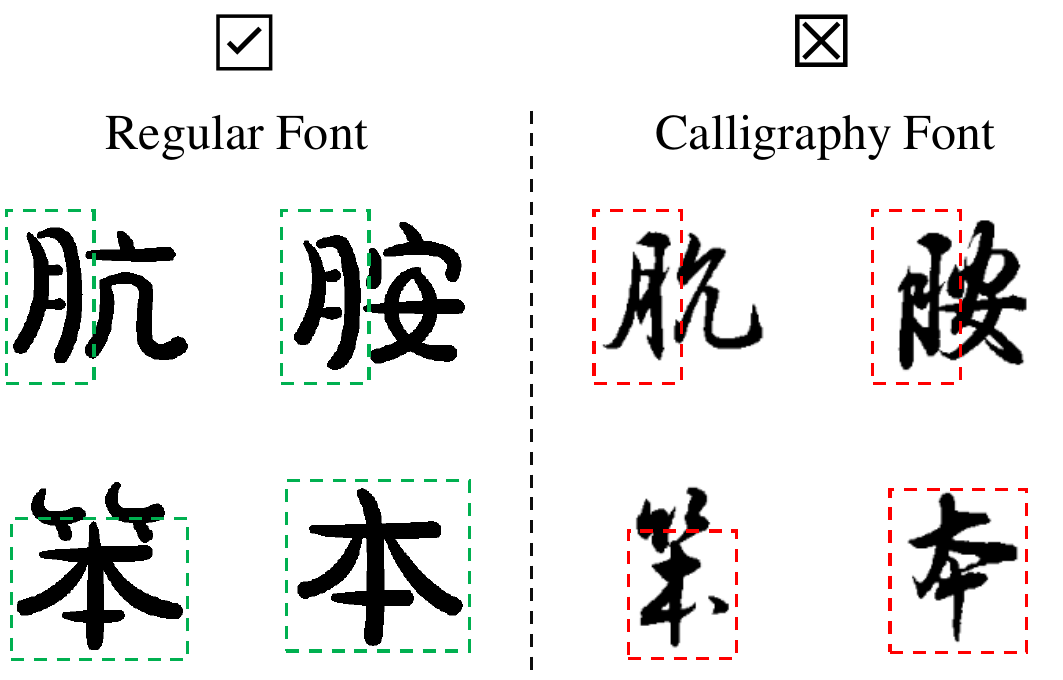}}
    \caption{Two assumptions of Chinese character prior knowledge. Regular Chinese fonts satisfy these assumptions, but calligraphy fonts don't.}
    \label{fig:assumption}
\end{figure}

As mentioned above, the task of few-shot Chinese font generation is challenging but of great practical value. There exist some works on this task, like EMD~\cite{zhang2018separating}, AGIS-Net~\cite{gao2019artistic}, etc., which concentrate on extracting and integrating features via well-designed loss functions or modules. Meanwhile, there also exist other methods that utilize prior knowledge of Chinese characters to extract structure-aware style representations~\cite{park2020few, park2021multiple, liu2022xmp} or encode prior knowledge into content labels~\cite{sun2017learning, wu2020calligan, huang2020rd}. In this manner, prior knowledge guided methods significantly improve the generation quality of most fonts and keep the fine-grained style consistency. Fig.~\ref{fig:assumption} shows two implicit consistency assumptions of such prior knowledge: 1) the same characters in different font styles have the same components and stroke order; 2) the same components in different characters but the same font style should be identical or similar. However, these methods often fail to satisfactorily handle calligraphy fonts with connected strokes. This is mainly due to the fact that calligraphy font designers tend to change the stroke order or simplify some strokes for better visual aesthetics, leading to calligraphy fonts conflicting with the component and stroke consistency. It can be observed from Fig.~\ref{fig:assumption} that glyphs in calligraphy styles are far more complex and hard to synthesize than those in regular font styles.

Compared to the prior knowledge, a writing trajectory (the sequence of key points) owns closer relationship and better consistency with the corresponding glyph image. Therefore, we try to use the writing trajectories instead of the prior knowledge and propose a dual-modality few-shot Chinese font generation model, DeepCalliFont, which processes and synthesizes Chinese characters represented in two modalities: glyph images and writing trajectories. More specifically, we map the dual-modality features to the same space using a distillation-restoration process and dual-modality representation learning. To ensure that the image and sequence branches generate consistent results, we propose the Image Feature Recombination module (IFR) and differentiable rasterization loss. Moreover, we also design a pre-training strategy using uni-modality data to relieve the lack of dual-modality data.

In summary, the contributions of this paper are fourfold:
\begin{itemize}
    \item We analyze the conflict between the consistency constraints implicit in the prior knowledge of Chinese characters and the aesthetic designing of calligraphy fonts. To resolve this conflict, we propose a novel few-shot Chinese font generation model, DeepCalliFont, effectively integrating the information of glyph images and writing trajectories via a dual-modality representation learning.
    \item We propose an Image Feature Recombination module and a differentiable rasterization loss to ensure the consistency of  two modalities, further improving the quality of synthesized glyph images.
    \item We design a pre-training strategy based on the coordinated representation of our DeepCalliFont, compensating for the lack of dual-modality data by exhaustively exploiting available uni-modality data.
    \item Quantitative and qualitative experiments demonstrate the superiority of our DeepCalliFont in the tasks of both regular and calligraphy font synthesis to existing methods, and the effectiveness of the proposed modules.
\end{itemize}

\section{Related Work}
\subsection{Glyph image synthesis}
Glyph image synthesis is a sub-task of image synthesis. Early attempts applied image generation methods to synthesize glyph images. For example, Tian~\cite{tian2017zi2zi} proposed zi2zi by modifying pix2pix~\cite{isola2017image}. Jiang et al.~\shortcite{jiang2017dcfont} improved zi2zi and designed DCFont, an end-to-end font style transfer system, to generate Chinese font with less online data. For the few-show font generation task, Azadi et al.~\shortcite{azadi2018multi} proposed an end-to-end stacked conditional GAN model, MC-GAN. Zhang et al.~\shortcite{zhang2018separating} used two encoders to extract content and style features, and applied skip-connections to ensure glyph correctness. Gao et al.~\shortcite{gao2019artistic} designed AGIS-Net with three discriminators to transfer both glyph shape and texture. There also exist some methods that utilize a large number of unpaired data to avoid time-consuming annotation. Wen et al.~\shortcite{wen2021zigan} designed an end-to-end Chinese calligraphy font generation framework, ZiGAN, utilizing unpaired data to catch fine-grained features. Xie et al.~\shortcite{xie2021dg} proposed an unsupervised method, DGFont, whose key idea is using deformable convolutions to properly transfer the source glyph features to the features of target glyph images.

To synthesize better glyph details, some researchers also sought the help of high-level prior knowledge of Chinese characters. For example, SA-VAE~\cite{sun2017learning} and CalliGAN~\cite{wu2020calligan} added the stroke or component encoding vectors into the feature. Zeng et al.~\shortcite{zeng2021strokegan}, Huang et al.~\cite{huang2020rd} and Kong et al.~\shortcite{kong2022look} designed stroke-aware or component-aware discriminators. Based on the same components in characters, Park et al. proposed LF-Font~\cite{park2020few} and MX-Font~\cite{park2021multiple}, establishing a relation between the local features and components. Similarly, Tang et al.~\shortcite{tang2022few} developed an algorithm to search more suitable reference sets based on the prior knowledge of the existence of repeating components in FS-Font. Although these existing methods can synthesize visually-pleasing glyph images, they typically fail to synthesize satisfactory calligraphy fonts, which do not meet the consistency assumptions in prior knowledge.
 
\subsection{Writing trajectory synthesis}
The writing trajectory synthesis task aims to generate the writing trajectories of glyphs in target font styles. Compared with glyph images, writing trajectory synthesis is more challenging. Because there is no brush rendering, the trajectory style is mainly reflected by global structures and connected strokes. Generally, writing trajectories represent as a sequence of key points with drawing control labels. Therefore, some researchers used sequential models (e.g., RNN and LSTM) to generate target font trajectories. For example, Ha~\shortcite{ha2015chinese} used RNN to synthesize writing trajectories. Zhang et al.~\shortcite{zhang2017drawing} also used RNN to identify and generate Chinese character trajectories and proposed a pre-processing method to make data suitable to sequential models. Tang et al.~\shortcite{tang2019fontrnn} proposed an RNN-based network with a monotonic attention mechanism. And later, they proposed WriteLikeYou~\cite{tang2021write} with two BiLSTM encoders and attention modules.

\begin{figure*}[htp]
    \centering
    \includegraphics[width=0.9\textwidth]{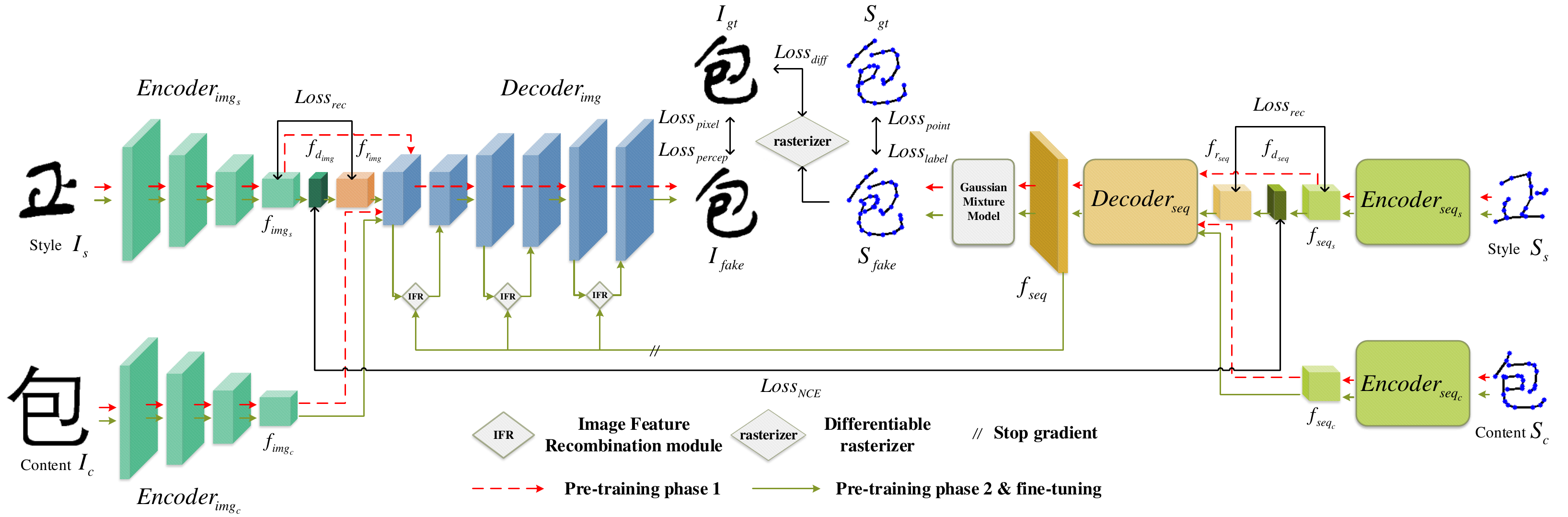}
    \caption{An overview of our model consisting of an image branch and a sequence branch. We first train two branches separately with uni-modality data (red dot lines) and then train the entire network jointly with dual-modality data (green solid lines).}
    \label{fig:overview}
\end{figure*}

On the other hand, there also exist some approaches that deform the source trajectory sequence to generate the target trajectory sequence. For example, Miyazaki et al.~\shortcite{miyazaki2019automatic} proposed a stroke extraction method and calculated the deformation matrix from the source skeleton to the target one. Liu and Lian~\cite{liu2021fontrl} proposed FontRL, using reinforcement learning to perform multiple TPS transformations on the stroke skeleton and synthesizing the glyph skeleton. FontRL performs satisfactorily but requires that the source skeleton has the same number of strokes as the target. Thereby, it is unsuited for synthesizing calligraphy fonts with omitted strokes.

A similar task is vector font synthesis~\cite{lopes2019learned, carlier2020deepsvg, smirnov2020deep, reddy2021im2vec, wang2021deepvecfont}, which synthesizes drawing commands instead of writing trajectories. These methods can output vector font libraries directly. However, it is still hard for existing methods to directly synthesis high-quality vector glyphs for complex Chinese characters~\cite{aoki2022svg, wang2023deepvecfont}. Therefore, we choose writing trajectories as the sequence modality in our work.

\subsection{Multi-modality font generation}
A Glyph can be represented in many modalities, such as the rasterization image, writing trajectory, glyph layout, components, etc. Some methods integrate two or more modalities to improve the synthesized font quality. For example, Wang et al.~\shortcite{wang2021deepvecfont} proposed DeepVecFont concatenating features of rasterization images and vector glyphs. SE-GAN~\cite{yuan2022se} used skeleton images to guide glyph image synthesis. Liu et al.~\shortcite{liu2022xmp} designed XMP-Font integrating rasterization images and stroke order with cross-attention modules. Unlike the above methods using the ``joint representation''~\cite{baltruvsaitis2018multimodal}, we choose the ``coordinated representation''~\cite{baltruvsaitis2018multimodal} because the glyph image contains information about other modalities. To be specific, we map dual-modality features to the same domain via a distillation-restoration process instead of concatenating them.

\section{Method}
As shown in Fig.~\ref{fig:overview}, DeepCalliFont is an end-to-end trainable network, consisting of two symmetrical sub-networks, the image branch and the sequence branch. In this section, we first introduce the data representation in our model. Then, we describe the image and sequence branches in detail, respectively, and present the dual-modality representation learning used to map features of two modalities into the same domain. Finally, we briefly introduce the pre-training strategy of our method.

\begin{figure}[tp]
    \centering
    \includegraphics[width=0.7\columnwidth]{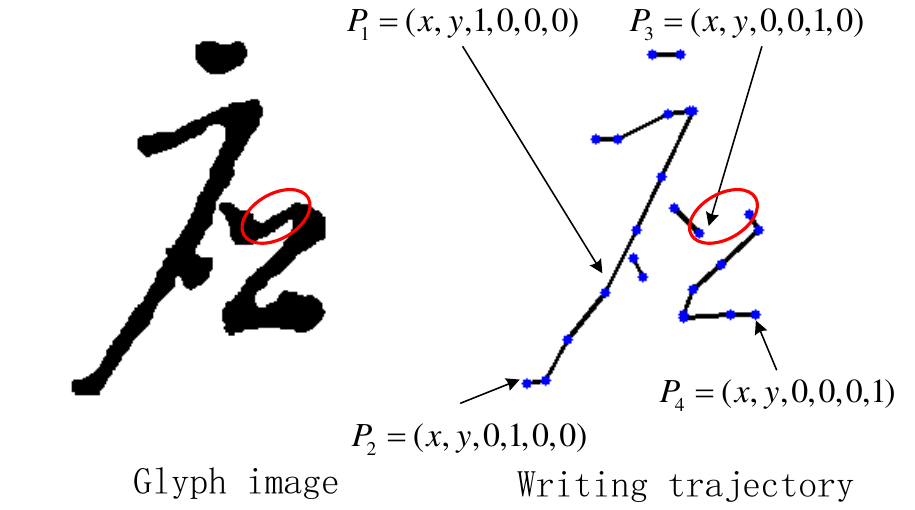}
    \caption{The two modalities used in DeepCalliFont: glyph images and writing trajectories. The red circle highlights a connected stroke in glyph image, so the format-6 representation of the corresponding point is $(x, y, 0, 0, 1, 0)$ (i.e., the control label $p_3=1$).}
    \label{fig:format6}
\end{figure}

\subsection{Data representation}
The proposed DeepCalliFont processes and synthesizes both glyph images ($128\times 128$) and the corresponding writing trajectories (see Fig.~\ref{fig:format6}). We denote the content, style, and target glyph images as $I_c$, $I_s$, and $I_{gt}$, respectively. For the sequence modality, we denote the writing trajectory of a character as a sequence of key points [$P_1$, $P_2$, $P_3$,...,$P_n$]. Some existing methods use format-3~\cite{graves2013generating} or format-5~\cite{ha2015chinese, tang2019fontrnn, tang2021write} to represent the key point $P_i$, which consists of a point coordinate and a control label. As shown in Fig.~\ref{fig:format6}, to distinguish whether there is a connected stroke between strokes or not, we extend the original format-5~\cite{tang2021write} into a new data format, format-6, which can be represented as $(x, y, p_1, p_2, p_3, p_4)$, where $x, y \in \mathbf{[-1,1]}$ denote the point coordinate values, and $(p_1, p_2, p_3, p_4)$ is a one-hot control label. $p_1 = 1$ means that there should be a visible line segment between this point and the next one; $p_2=1$ denotes the end of a stroke with no connection to the next stroke; $p_3=1$ means the end of a stroke that has a visual connection with the next stroke; and $p_4 = 1$ means the end of the writing process.\label{sec:format-6}

\subsection{Sequence branch}
The sequence branch is a transformer-based seq2seq model, which can be formulated as:
\begin{equation}
\begin{aligned}
    & f_{{seq}_s} = \frac{1}{L}\sum^{L}_{i=1}Encoder_{{seq}_s}(S_s)^i; \\
    & f_{{seq}_c} = Encoder_{{seq}_c}(S_c); \\
    & f_{seq} = Decoder_{seq}(f_{{seq}_c}, f_{seq_s}); \\
    & S_{fake-point} = \mathit{GMM}(f_{seq}); \\
    & S_{fake-label} = ControlLabelClassifier(f_{seq});
    \label{eq:seq}
\end{aligned}
\end{equation}
where $S_s$ and $S_c$ denote the style and content reference sequences, respectively, and $L$ is the sequence length.

Following existing works, we also use Gaussian Mixture Model (GMM)~\cite{tang2019fontrnn, tang2021write} to model the coordinates of key points as M bivariate normal distributions, and predict the control label by a control label classifier, which is a fully connected layer. The points and control labels can be represented as:
\begin{align}
    & S_{fake-point} = \{\pi^i, \mu_x^i, \mu_y^i, \sigma_x^i, \sigma_y^i, \rho^i_{xy}\}^M_{i=1};  \\
    & S_{fake-label} = [q_1, q_2, q_3, q_4];
\end{align}
where $\pi$ denotes the probability of M distributions, $q$ denotes the predicted control labels, and $M =20$ by default.

The losses of the GMM and control label classifier are defined as:
\begin{align}
    & loss_{point} = -\frac{1}{L}\sum_{i=1}^{L}log(p(x, y)); \\
    & p(x, y) = \sum_{i=1}^{M}\pi^i\mathcal{N}(x, y|\mu_x^i, \mu_y^i, \sigma_x^i, \sigma_y^i, \rho^i_{xy}); \\
    & loss_{label} = CE(q, label_{control});
\end{align}
where $CE$ means the cross entropy loss, and $label_{control}$ denotes the target control label.

\textbf{Differentiable rasterizer loss.} To generate the writing trajectory located in the target glyph image, we proposed a differentiable rasterization loss based on Unsigned Distance Field (UDF). UDF defines a distance field between a grid and an object. The object in the sequence branch is a writing trajectory, which is represented as the set of all line segments between the key points. So the UDF in DeepCalliFont can be formalized as follows:
\begin{align}
    &Dist(x, L) = \underset{l\in L}{min}(Dist(x, l)),\quad x \in grid; \\
    &grid = \{(i, j)~|~i \in \{0, 1,...,H-1\}, j \in \{0, 1,...,W-1\}\}; \\
    &Dist(x, l) = \left \{
    \begin{array}{ll}
        |\vec{x p_s}|, & \vec{p_s p_t} \cdot \vec{p_s x} < 0, \\
        |\vec{x p_t}|, & \vec{p_t p_s} \cdot \vec{p_t x} < 0, \\
        \frac{abs(\vec{x p_s} \times \vec{x p_t})}{|\vec{p_s p_t}|}, & otherwise;
    \end{array}
    \right.
\end{align}
where $L$ denotes the writing trajectory, $p_s$ and $p_t$ are the starting and ending points of the line segment $l$. Moreover, the distance $Dist(x, l)$ is set to $\infty$ if the line $l$ is invisible (i.e., $p_1=0$ in format-6).

After getting the UDF of the synthesized writing trajectory, we render it into a raster image. To ensure the differentiability of the rendering process, we use the $Sigmoid$ function to quantify the pixel value and get the image $I_\mathit{diff}$. The rendering process can be formalized as follows:
\begin{align}
\begin{array}{r}
    I_\mathit{diff}= \{1 - Sigmoid(\theta (Dist(x, S_{fake-point})-w))| \\
    x \in grid\},
\end{array}
\label{for:diff}
\end{align}
where $w$ denotes the line width, and $\theta$ is a hyper-parameter. In this paper, $\theta$ and $w$ are chosen as 100 and 2, respectively.

Then, we design a loose differentiable rasterizer loss, which is defined as:
\begin{align}
    loss_\mathit{diff} = ||ReLU(I_\mathit{diff}- I_{gt})||_2^2.
\end{align}

\textbf{Loss function.} Finally, we calculate the complete loss function of the sequence branch by:
\begin{align}
    loss_{seq} = \lambda_1 loss_{point} + \lambda_2 loss_{label} + \lambda_3 loss_\mathit{diff},
\end{align}
where $\lambda_1$, $\lambda_2$, and $\lambda_3$ are hyper-parameters that control the weights of $loss_{point}$, $loss_{label}$ and $loss_\mathit{diff}$.

\subsection{Image branch}
The image branch is a CNN-based auto-encoder, including a style encoder, a content encoder, and a glyph decoder with skip connections between the content encoder and the decoder. The image branch can be formulated as:
\begin{equation}
\begin{aligned}
    & f_{{img}_s} = Encoder_{{img}_s}(I_s); \\
    & f_{{img}_c} = Encoder_{{img}_c}(I_c); \\
    & I_{fake} = Decoder_{img}(f_{{img}_s}, f_{{img}_c});
\end{aligned}
\end{equation}
where $I_s$ and $I_c$ denote the style and content reference images, respectively.

\textbf{Image Feature Recombination module.} Since the image feature aims to give a global picture of the glyph while the sequence feature captures information of local writing details, it is possible to integrate these two complementary features to create a more discriminative one. Therefore, we design the Image Feature Recombination (IFR) module to recombine the image feature under the guidance of the sequence feature through an attention module. Because image features and sequence features are in different distributions (i.e., $f_{img}\sim p(img)$ and $f_{seq}\sim p(seq)$), we utilize two attention modules to recombine the image feature instead of directly concatenating these two types of features in the IFR module. In this manner, we can keep the recombined image feature in the same domain as the original image feature.\label{sec:IFRmodule}

\begin{figure}[tp]
    \centering
    \includegraphics[width=0.9\columnwidth]{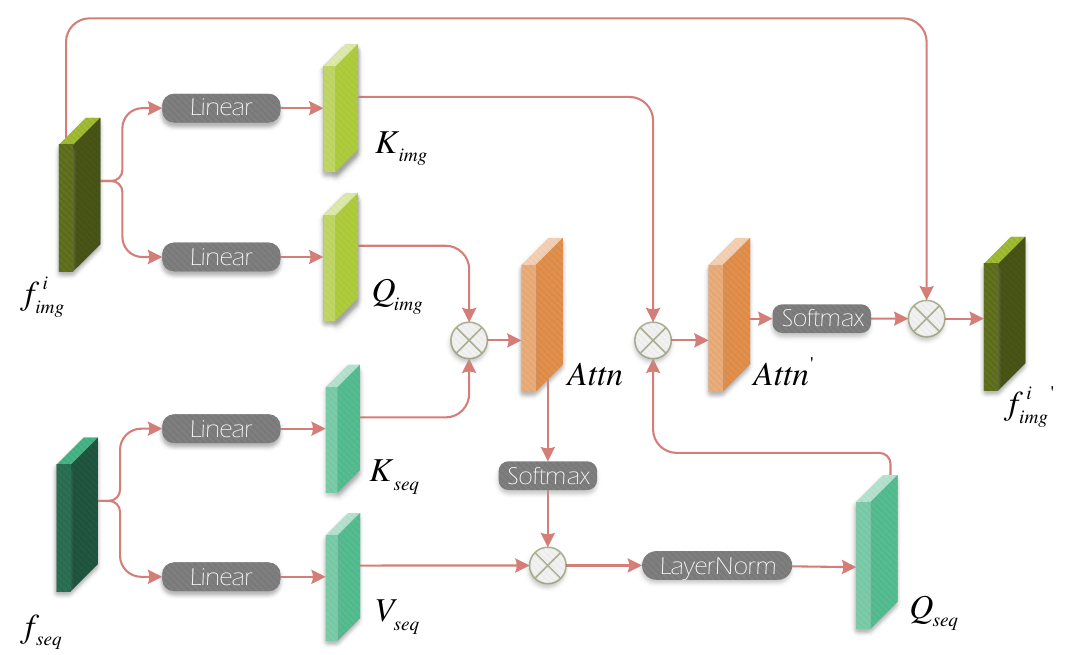}
    \caption{The illustration of the proposed Image Feature Recombination module. To keep the original feature $f_{img}^{i}$ and the recombined feature ${f_{img}^{i}}^{'}$ in the same domain, we design a attention module mapping the feature from image to sequence domain and another attention module mapping the feature from sequence to image domain.}
    \label{fig:IFRmodule}
\end{figure}

As shown in Fig.~\ref{fig:IFRmodule}, we first map the image and sequence features to query (Q), key (K), and value (V) vectors through linear layers:
\begin{equation}
\begin{aligned}
    & Q_{img} = Linear_{Q_{img}}(f_{img}^{i})^{T} \in R^{(H\times W)\times d}; \\
    & K_{img} = Linear_{K_{img}}(f_{img}^{i})^{T}  \in R^{(H\times W)\times d}; \\
    & K_{seq} = Linear_{K_{seq}}(f_{seq}) \in R^{L\times d};\\
    & V_{seq} = Linear_{V_{seq}}(f_{seq}) \in R^{L\times d}; 
\end{aligned}
\end{equation}
where $f_{img}^{i}$ denotes the feature outputted from the $ith$ layer of the image decoder, $f_{seq}$ denotes the feature generated by the sequence decoder, $H\times W$ is the size of image feature, $L$ is the length of sequence feature, and $d$ is the feature size.

Then, we calculate $Q_{seq}$ through an attention module and layer normalization:
\begin{equation}
    Q_{seq} = LayerNorm(softmax(\frac{Q_{img}K^T_{seq}}{\sqrt{d}})V_{seq}).
    \label{eq:attn1}
\end{equation}

After getting the feature $Q_{seq}$, we recombine the original image feature $f_{img}^{i}$ by another attention module:
\begin{equation}
    {f_{img}^{i}}^{'} = softmax(\frac{Q_{seq}K^T_{img}}{\sqrt{d}})f_{img}^{i}.
    \label{eq:attn2}
\end{equation}

Under the guidance of the sequence feature, we get a new feature ${f_{img}^{i}}^{'}$ by recombining image features $f_{img}^{i}$ extracted from different layers. We assume that the pixels in an image are i.i.d. and the convolution layer uses a fixed kernel to extract the feature from the image, so the features are also i.i.d. in the spatial level. Moreover, the attention mechanism is a linear combination of the value vector. Therefore, the IFR module keeps the recombined image feature ${f_{img}^{i}}^{'}$ in the same domain as the original image feature $f_{img}^{i}$.

In addition, we stop the gradient of sequence feature $f_{seq}$ to avoid that the unconverged image branch in the early training phase negatively influences the sequence branch.

\textbf{Loss function.} We use MAELoss and the perceptual loss~\cite{johnson2016perceptual} to optimize the image branch:
\begin{equation}
\begin{aligned}
    & loss_{pixel} = |I_{gt}-I_{fake}|; \\
    & loss_{percep} = |Gram(vgg(I_{gt}))-Gram(vgg(I_{fake}))|; \\
    & loss_{img} = \lambda_4 loss_{pixel} + \lambda_5 loss_{percep}; 
\end{aligned}
\end{equation}
where $vgg$ denotes a pre-trained VGG network, $Gram$ means the Gram matrix of features, $\lambda_4$ and $\lambda_5$ are hyper-parameters that control the weights of two losses.

\subsection{Dual-modality representation learning}
As above mentioned, image and sequence features correspond to different feature domains. Therefore, we map style features $f_{img_s}$ and $f_{seq_s}$ into the same domain through \textbf{distillation} (dimension reduction) and \textbf{restoration} (dimension increasement). To be specific, we first distillate the feature by a linear layer and normalize it. Then we restore the feature via another linear layer to reconstruct the original feature. The distillation-restoration process can be formulated as:
\begin{equation}
\begin{aligned}
    & f_{d_x} = \frac{Linear_{d}(f_{x_s})}{|Linear_{d}(f_{x_s})|},\quad x \in \{img, seq\}; \\
    & f_{r_x} = Linear_{r}(f_{d_x}),\quad x \in \{img, seq\};
\end{aligned}
\end{equation}
where the features $f_{x_s}$ and $f_{r_x}$ have the same dimension, and the dimension of $f_{d_x}$ is the half of theirs.

After getting the distilled features $f_{d_x}$, we apply the InfoNCE loss~\cite{oord2018representation} to the distilled image and sequence features. Because the feature vectors are normalized, we can measure the similarity with the cosine distance. The InfoNCE loss is computed by:
\begin{equation}
\begin{aligned}
    & sim(img_i, seq_j) = cos(f_{d_{img_i}}, f_{d_{seq_j}}); \\
    & loss_{\mathit{NCE}} = \frac{1}{B}\sum^{B}_{i=1}-log\frac{exp(\tau \cdot sim(img_i, seq_i))}{\sum^{B}_{j=1}exp(\tau \cdot sim(img_i, seq_j))};
\end{aligned}
\label{eq:nce}
\end{equation}
where $f_{d_x}$ denotes the distilled feature, $\tau$ denotes the learnable temperature, and $B$ means the batch size.

After restoration, we feed $f_{r_x}$ into the decoder instead of $f_{x_s}$. To keep the consistency between the reconstructed feature $f_{r_x}$ and the original $f_x$, we also adopt a feature reconstruction loss:
\begin{equation}
\begin{aligned}
    loss_{rec} = ||f_{img_s} - f_{r_{img}}||^2_2+||f_{seq_s} - f_{r_{seq}}||^2_2.
\end{aligned}
\end{equation}
Moreover, we apply deep metric learning~\cite{aoki2021few} to the features $f_{x_s}$ and $f_{d_x}$, which can be formalized as:
\begin{equation}
\begin{aligned}
    & loss_{dml} = \\
    & CE(D_1(f_{img_s}), label) + CE(D_2(f_{d_{img}}), label) + \\
    & CE(D_3(f_{seq_s}), label) + CE(D_4(f_{d_{seq}}), label), \\
\end{aligned}
\label{eq:dml}
\end{equation}
where $D$ denotes the style classifier (a fully connected layer), and $label$ denotes the ground-truth font style.

We map the image and sequence features into the same domain through distillation and restoration. In this manner, the recombined feature can contain dual-modality information. In summary, the complete loss function of our DeepCalliFont is defined as:
\begin{equation}
\begin{aligned}
    loss =\, & loss_{img} + loss_{seq} + \\
    & loss_{\mathit{NCE}} + loss_{rec} + loss_{dml}.
\end{aligned}
\end{equation}

\subsection{Pre-training strategy}
Considering that it is hard to build large-scale dual-modality datasets, we design a two-phase pre-training strategy. As shown in Fig.~\ref{fig:overview}, in phase 1, we train the image and sequence branches separately using a large amount of uni-modality data. In phase 2, we add modality interaction modules, including the IFR module, the differentiable rasterizer loss, and the distillation-restoration module, and jointly train two branches using a small amount of dual-modality data.

\begin{figure}[tp]
    \centering
    \includegraphics[width=0.9\columnwidth]{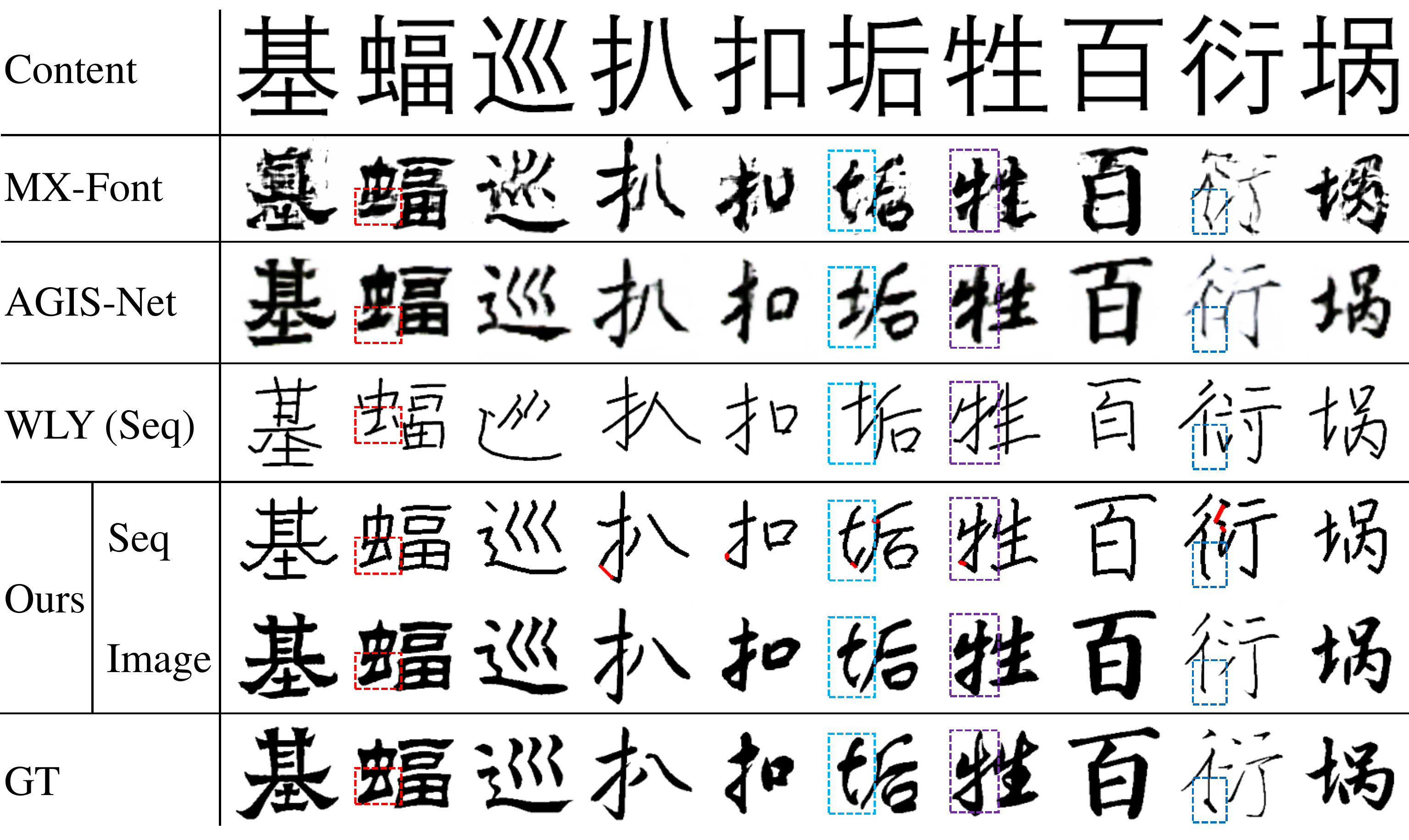}
    \caption{Comparison of calligraphy font synthesis results obtained by DeepCalliFont and other three methods. For the proposed DeepCalliFont, we show the synthesis results of two branches. We highlight the synthesis results of connected strokes with boxes. Moreover, the red lines in synthesized sequences denote the connected strokes.}
    \label{fig:calli_show}
\end{figure}

\section{Experiments}
\subsection{Experimental settings}
We use 251 fonts and CASIA Online Chinese Handwriting Databases~\cite{liu2011casia} to pre-train two branches separately in the pre-training phase 1. Then, in the pre-training phase 2, we selected 42 fonts used in~\cite{jiang2019scfont}, each of which consists of 3,000 glyph images and their corresponding writing trajectories, to train the whole model. Since there is no connected stroke annotation available in the above-mentioned datasets, we automatically generate pseudo annotations as follows. We define two strokes as connected if and only if the distance between the previous stroke's endpoint and the next stroke's start point is less than a given threshold and the distance between corresponding points in the mean skeleton is greater than the threshold.

To demonstrate the generalizability of the proposed DeepCalliFont, we construct two test datasets, i.e., a regular font test set (without connected strokes) and a calligraphy font test set (with connected strokes). We use 30 fonts in the dataset collected by Liu and Lian~\cite{liu2023fonttransformer} as the regular font test set. We also collect ten calligraphy fonts as the other test set. In our experiments, we fine-tune networks on 100 sample characters and test them on other 6,663 Chinese characters. Then, we evaluate the quality of synthesized glyph images based on MAE, FID~\cite{heusel2017gans}, and LPIPS~\cite{zhang2018perceptual}, and calculate the similarity between generated and ground-truth sequences via dynamic time warping (DTW)~\cite{bahdanau2014neural}.

\subsection{Quantitative and qualitative comparison}
We compare the performance of our DeepCalliFont with seven existing methods, including zi2zi~\cite{tian2017zi2zi}, EMD~\cite{zhang2018separating}, AGIS-Net~\cite{gao2019artistic}, MX-Font~\cite{park2021multiple}, DG-Font~\cite{xie2021dg}, FontRL~\cite{liu2021fontrl}, and WriteLikeYou (WLY)~\cite{tang2021write}. Since zi2zi, EMD, DGFont perform poorly, and FontRL requires the same number of strokes, we compare our method with the others on calligraphy fonts. For the sake of fair comparison, we pre-train and fine-tune all networks with the same data, including uni-modality and dual-modality data.

\begin{figure}[tp]
    \centering
    \includegraphics[width=0.9\columnwidth]{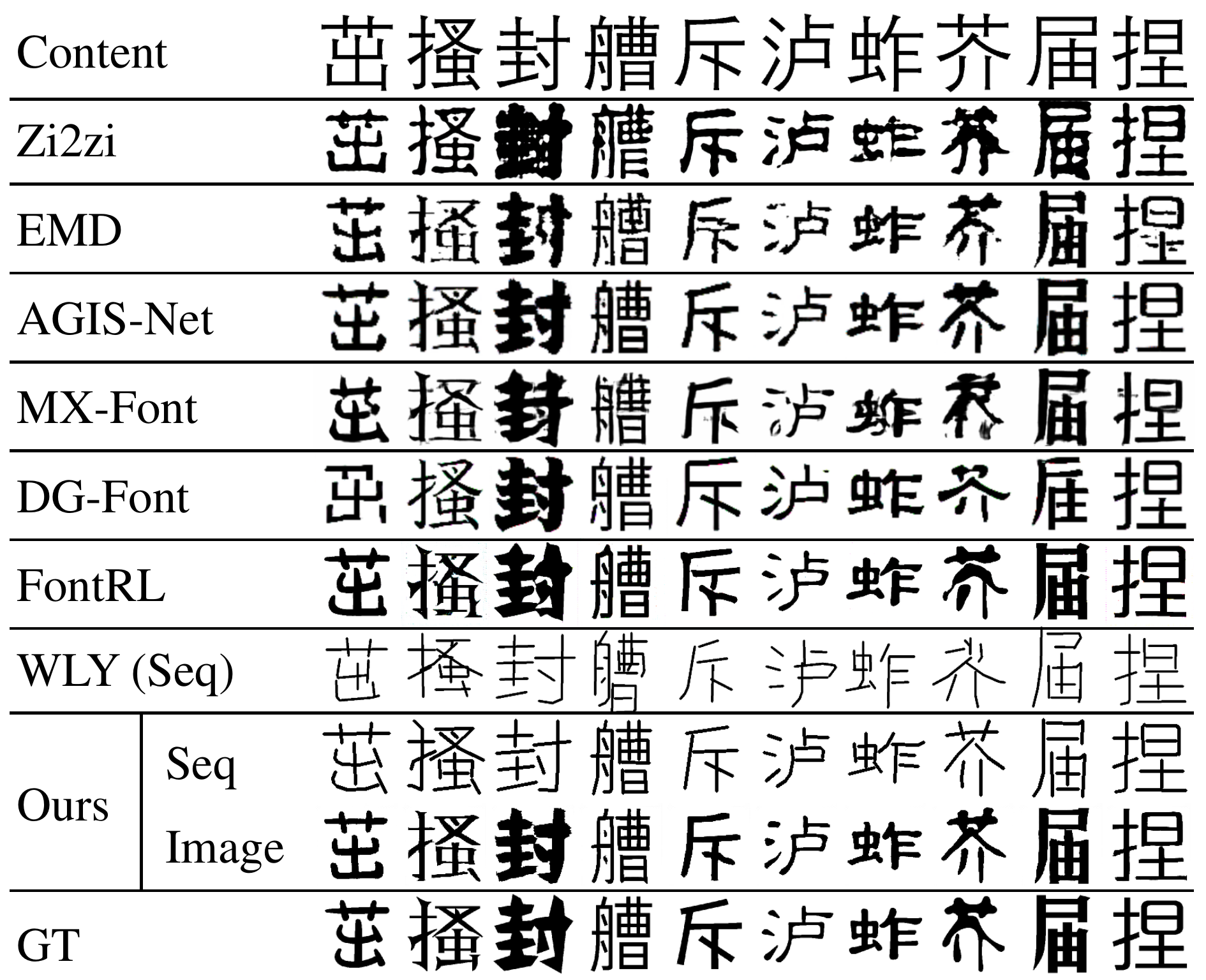}
    \caption{Comparison of regular font synthesis results generated by DeepCalliFont and other methods.}
    \label{fig:standard_show}
\end{figure}

Tab.~\ref{tab:quantitative} shows that our DeepCalliFont outperforms other methods in all metrics. The proposed approach exhibits a significant advantage against them, especially in the metrics of FID and LPIPS. As shown in Fig.~\ref{fig:calli_show}, MX-Font and AGIS-Net struggle to synthesize glyphs with connected strokes, although they use adversarial learning or additional prior knowledge. WriteLikeYou solves this problem to some extent, but it often fails to satisfactorily reconstruct the whole glyph structure. On the contrary, our DeepCalliFont captures the difference on strokes between the source font and the target font, and synthesizes the highest-quality glyph images in calligraphy styles. Since our model imposes several constraints to keep the features of two modalities consistent, it can synthesize consistent glyph images and writing trajectories that coincide well. 

\begin{table}[tp]
    \centering
    \begin{tabularx}{\columnwidth}{l|MMM}
        \hline
        \multicolumn{4}{c}{Calligraphy Fonts} \\
        \hline
        Method & MAE$\downarrow$ & FID$\downarrow$ & LPIPS$\downarrow$ \\
        \hline
        AGIS-Net& 0.1383 & 463.9 & 0.2857\\
        \hline
        MX-Font & 0.1850 & 235.1 & 0.2754 \\
        \hline
        WriteLikeYou & 0.1757 & 167.2 & 0.2714 \\
        \hline
        ours & \textbf{0.1254} & \textbf{14.6} & \textbf{0.1718} \\
        \hline
    \end{tabularx}
    \begin{tabularx}{\columnwidth}{l|MMMM}
        \multicolumn{5}{c}{Regular Fonts} \\
        \hline
        Method & MAE$\downarrow$ & FID$\downarrow$ & LPIPS$\downarrow$ & DTW$\downarrow$ \\ 
        \hline
        zi2zi & 0.1719 & 137.1 & 0.4050 & - \\
        \hline
        EMD & 0.1538 & 184.5 & 0.4628 & - \\
        \hline
        AGIS-Net & 0.1354 & 157.7 & 0.4584 & - \\
        \hline
        MX-Font & 0.2115 & 137.4 & 0.5055 & - \\
        \hline
        DG-Font & 0.1886 & 150.2 & 0.4878 & - \\
        \hline
        FontRL & 0.1350 & \textbf{40.7} & 0.2588 & 15.304\\
        \hline
        WriteLikeYou & 0.2223 & 114.4 & 0.4565 & 5.920 \\
        \hline
        ours & \textbf{0.1242} & 105.9 & \textbf{0.1837} & \textbf{1.347} \\
        \hline
    \end{tabularx}
    \caption{Quantitative results of our DeepCalliFont and other methods on calligraphy and regular fonts.}
    \label{tab:quantitative}
\end{table}

Although DeepCalliFont is specifically designed to handle calligraphy fonts consisting of glyphs that often contain connections between strokes, its performance on regular fonts is also satisfactory. Synthesis results of different methods are shown in Fig.~\ref{fig:standard_show}, from which we can see that our DeepCalliFont still outperforms other state-of-the-art methods in the task of regular font synthesis.

\begin{table*}[tp]
    \centering
    \begin{tabular}{cccccccc|ccc}
    \hline
    Image & Seq & IFR & Distillation & Diff & NCE Loss & DML & Uni-modality pre-train & MAE$\downarrow$ & FID$\downarrow$ & LPIPS$\downarrow$ \\
    \hline
    \checkmark & \checkmark & \checkmark & \checkmark & \checkmark & \checkmark & \checkmark & \checkmark & \textbf{0.1254} & \textbf{14.6} & \textbf{0.1717} \\ 
    \hline
    \checkmark & \checkmark &  & \checkmark & \checkmark & \checkmark & \checkmark & \checkmark & 0.1270 & 23.7 & 0.1882 \\
    \hline
    \checkmark & \checkmark & \checkmark &  & \checkmark & \checkmark & \checkmark & \checkmark & 0.1311 & 24.6 &  0.1969 \\
    \hline
    \checkmark & \checkmark & \checkmark & \checkmark &  & \checkmark & \checkmark & \checkmark & 0.1285 & 24.6 & 0.1925 \\
    \hline
    \checkmark & \checkmark & \checkmark & & \checkmark &  & \checkmark & \checkmark & 0.1279 & 27.0 & 0.1961 \\
    \hline
    \checkmark & \checkmark & \checkmark & \checkmark & \checkmark & \checkmark & & \checkmark & 0.1300 & 25.4 & 0.1943 \\
    \hline
    \checkmark & & & & & & \checkmark & \checkmark & 0.1348 & 21.4 & 0.1821 \\
    \hline
    \checkmark & \checkmark & \checkmark & \checkmark & \checkmark & \checkmark & \checkmark & & 0.1355 & 43.4 & 0.2197 \\
    \hline
    \end{tabular}
    \caption{Ablation study results. Image: image branch; Seq: sequence branch; IFR: Image Feature Recombination module; Distillation: ``Distillation-restoration''
    module; Diff: differentiable rasterization loss; NCE Loss: the InfoNCE loss (Eq.~\ref{eq:nce}); DML: Deep metric learning (Eq.~\ref{eq:dml}); Uni-modality pre-train: the pre-training stage 1.}
    \label{tab:ablation}
\end{table*} 

\begin{figure}[htp]
    \centering
    \includegraphics[width=0.9\columnwidth]{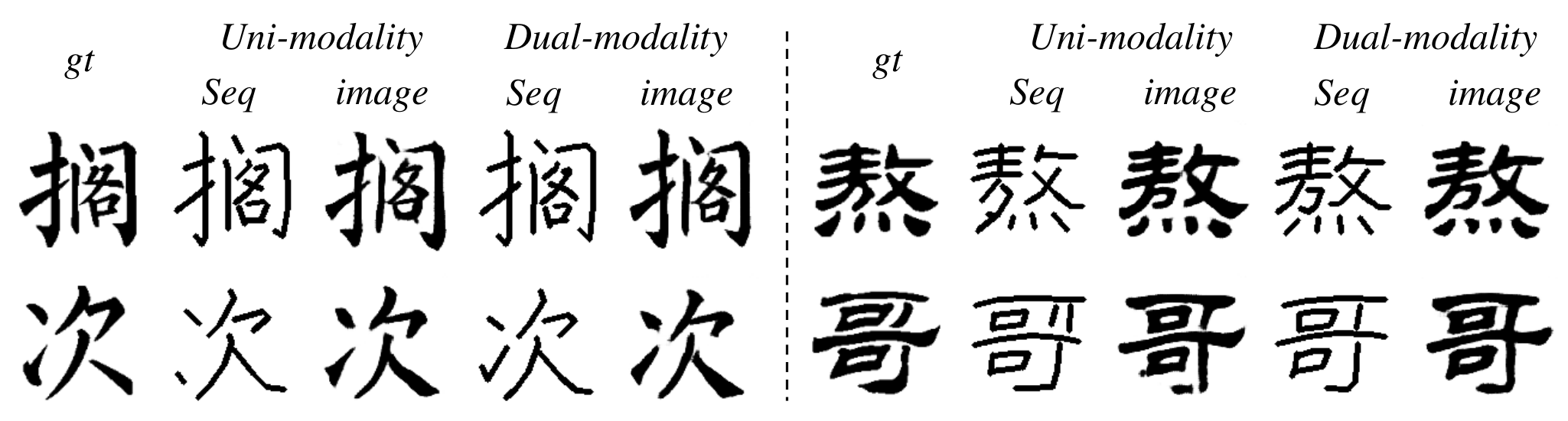}
    \caption{Comparison of synthesis results obtained by the uni-modality branch or the dual-modality model. The results of our dual-modality model have better glyph consistency and better details.}
    \label{fig:dual-modality}
\end{figure}

\subsection{Ablation studies}
In this subsection, we analyze the role of each proposed module or loss function. Tab.~\ref{tab:ablation} shows the quantitative results demonstrating that those proposed modules and loss functions all help to improve the performance of our method.

\textbf{Uni-modality vs dual-modality synthesis.} Tab.~\ref{tab:ablation} shows that our full model performs better in all metrics than the model using only the image branch. As we can see from Fig.~\ref{fig:dual-modality}, with the help of dual-modality representation learning and other proposed modules, our model is capable of synthesizing better glyph details.

\begin{figure}[tp]
    \centering
    \includegraphics[width=0.9\columnwidth]{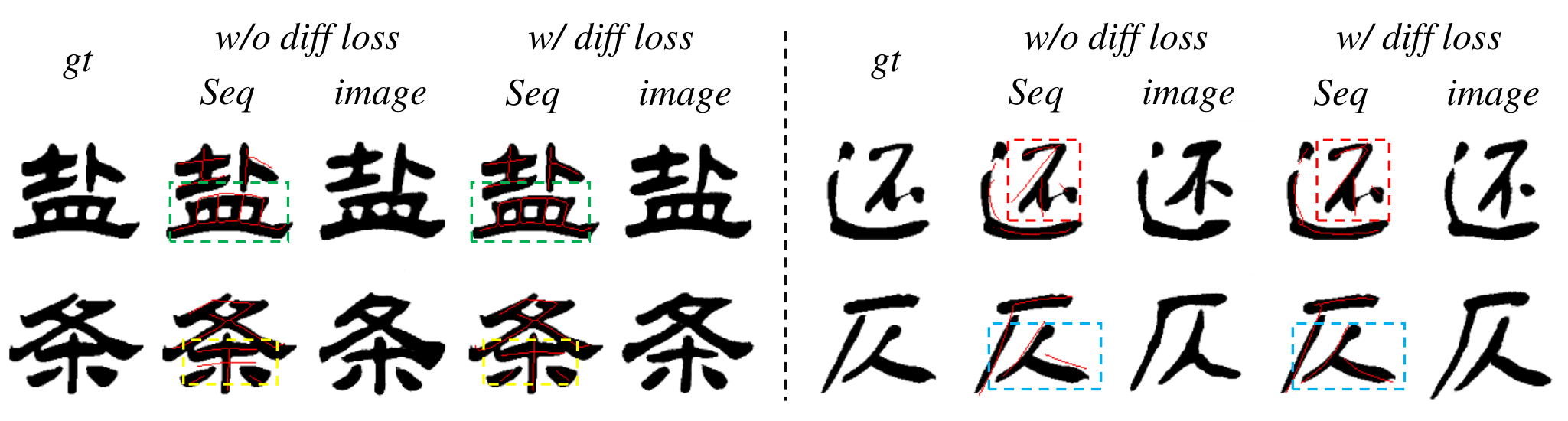}
    \caption{Comparison of synthesis results of our method with or without the differentiable rasterizer loss, which limits the generated trajectory inside the corresponding glyph.}
    \label{fig:rasterizer}
\end{figure}

\textbf{Differentiable rasterization loss.} The differentiable rasterization loss $loss_\mathit{diff}$ constrains the writing trajectories to be located inside the corresponding glyphs. From Fig.~\ref{fig:rasterizer} where the synthesized trajectories are drawn on the target glyph images, we can see that the writing trajectory synthesized with $loss_\mathit{diff}$ matches better with the target image.

\begin{figure}[tp]
    \centering
    \includegraphics[width=0.9\columnwidth]{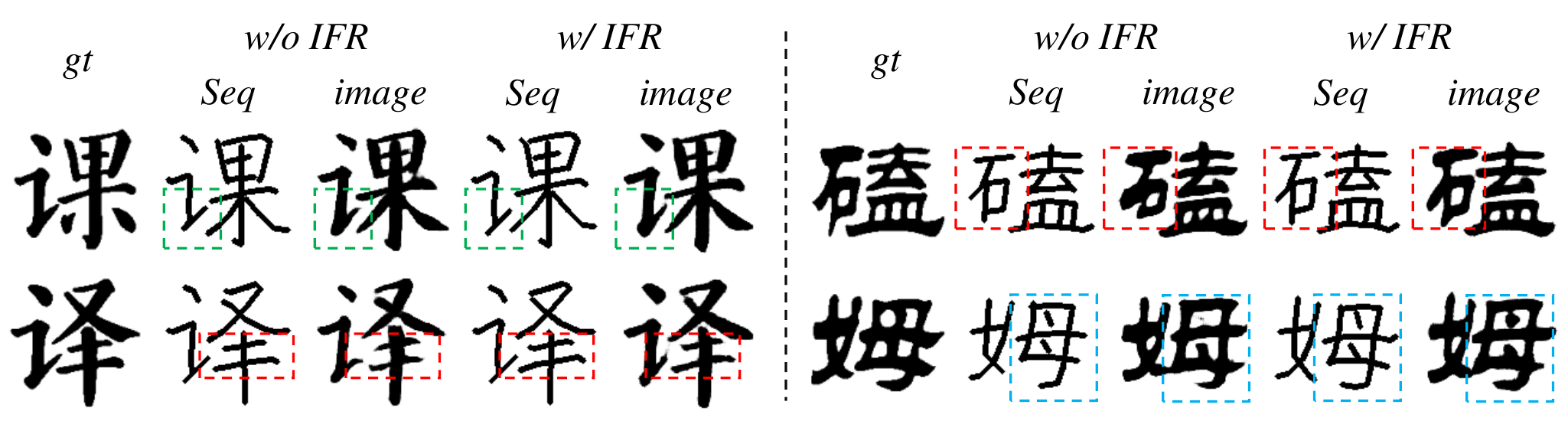}
    \caption{Comparison of synthesis results obtained by our method with or without the proposed IFR module. Through the IFR module, the image branch improves the details of strokes under the guidance of sequence features.}
    \label{fig:IFR_show}
\end{figure}

\textbf{Image Feature Recombination module.} The Image Feature Recombination (IFR) module aims to improve the result of the image branch with the help of sequence features. Empirically, the sequence branch converges faster and mostly generates correct glyph structures. We can see from Fig.~\ref{fig:IFR_show} that the image branch with IFR generates glyph images more consistent with the synthesized writing trajectories than the one without the IFR module. Moreover, the sequence feature can guide image branches to generate better details, which are difficult to correct through loss functions.

\textbf{``Distillation-restoration'' module.} Due to the above-mentioned pre-training strategy, the two branches were designed to be independent. Therefore, we design the ``distillation-restoration'' module to combine these two branches into a unified network. To verify the effectiveness of this module, we remove it and apply the InfoNCE loss directly on features $f_{img_s}$ and $f_{seq_s}$ (see Tab.~\ref{tab:ablation}).

\textbf{Deep metric learning.} Following Aoki et al.~\shortcite{aoki2021few}, we add deep metric learning into DeepCalliFont, helping our model distinguish font styles. Therefore, the $loss_{\mathit{dml}}$ mainly improves the method performance in of FID and LPIPS.

\textbf{Two-stage pre-training strategy} In Tab.~\ref{tab:ablation}, we compare the results obtained with and without the first stage pre-training, validating its effectiveness.

\subsection{Limitation}
Since the few-shot font generation task is essentially ill-posed, synthesized images may differ from the target ground truth but are visually pleasing. More discussions and failure cases can be found in the supplementary material.

\section{Conclusion}
In this paper, we proposed a novel end-to-end few-shot Chinese font synthesis method, DeepCalliFont. To handle the connected strokes frequently contained in calligraphy fonts, we designed a dual-modality (i.e., glyph images and writing trajectories) few-shot font generation network. Specifically, we utilized contrastive learning to keep the consistency of image and sequence branches at the feature level. We also proposed the Image Feature Recombination module and differentiable rasterization loss to generate consistent and high-quality results. Meanwhile, we designed a pre-training strategy to use a large amount of uni-modality data to compensate for the lack of dual-modality data. Both quantitative and qualitative experiments showed that the proposed DeepCalliFont can synthesize higher-quality calligraphy fonts with more reasonable connected strokes than other SOTA methods and also perform satisfactorily on regular fonts.

\section{Acknowledgments}
This work was supported by National Natural Science Foundation of China (Grant No.: 62372015), Center For Chinese Font Design and Research, and Key Laboratory of Intelligent Press Media Technology.

\bibliography{refer}

\end{document}